\documentclass[letterpaper]{article} 
\usepackage{aaai2026}  
\usepackage{times}  
\usepackage{helvet}  
\usepackage{courier}  
\usepackage[hyphens]{url}  
\usepackage{graphicx} 
\usepackage{natbib}  
\usepackage{caption} 
\usepackage{amssymb}
\usepackage{amsmath}
\usepackage{booktabs}
\usepackage{makecell}
\usepackage{multirow}
\usepackage{pifont}

\usepackage{algorithm}
\usepackage{algorithmic}

\usepackage{newfloat}
\usepackage{listings}
\DeclareCaptionStyle{ruled}{labelfont=normalfont,labelsep=colon,strut=off} 
\floatstyle{ruled}
\newfloat{listing}{tb}{lst}{}
\floatname{listing}{Listing}
\title{Generalized Geometry Encoding Volume for Real-time Stereo Matching}
\author{
    Jiaxin Liu, 
    Gangwei Xu, 
    Xianqi Wang, 
    Chengliang Zhang, 
    Xin Yang\thanks{Corresponding author.}
}

\affiliations{
    Huazhong University of Science and Technology\\
    \{ljx123, gwxu, xianqiw, z\_chengliang, xinyang2014\}@hust.edu.cn

}

\usepackage{bibentry}

\begin{document}

\maketitle

\begin{abstract}
Real-time stereo matching methods primarily focus on enhancing in-domain performance but often overlook the critical importance of generalization in real-world applications.
In contrast, recent stereo foundation models leverage monocular foundation models (MFMs) to improve generalization, but typically suffer from substantial inference latency.
To address this trade-off, we propose Generalized Geometry Encoding Volume (GGEV), a novel real-time stereo matching network that achieves strong generalization.
We first extract depth-aware features that encode domain-invariant structural priors as guidance for cost aggregation.
Subsequently, we introduce a Depth-aware Dynamic Cost Aggregation (DDCA) module that adaptively incorporates these priors into each disparity hypothesis, effectively enhancing fragile matching relationships in unseen scenes.
Both steps are lightweight and complementary, leading to the construction of a generalized geometry encoding volume with strong generalization capability.
Experimental results demonstrate that our GGEV surpasses all existing real-time methods in zero-shot generalization capability, and achieves state-of-the-art performance on the KITTI 2012, KITTI 2015, and ETH3D benchmarks.
Code: https://github.com/JiaxinLiu-A/GGEV.
\end{abstract}


\section{Introduction}
Stereo matching aims to estimate dense, pixel-wise disparity maps from a pair of rectified stereo images.
As a long-standing and challenging task in computer vision, it plays a fundamental role in a wide range of applications, including 3D reconstruction~\cite{liang2025parameter}, autonomous driving~\cite{li2025recogdrive,liang2025sood++}, and robotic navigation.
These real-world scenarios impose strict demands on both generalization and inference latency.

\begin{figure}[t]
\centering
\includegraphics[width=0.98\columnwidth]{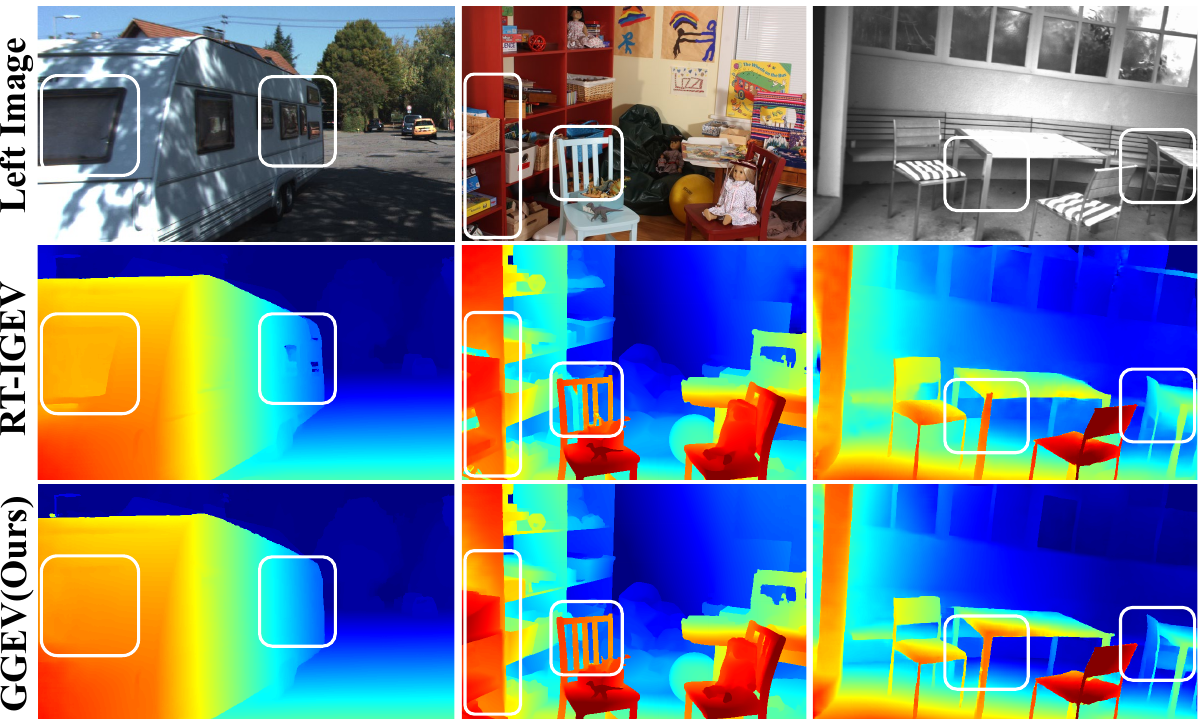} 
\caption{Zero-shot generalization comparison. All models are trained on Scene Flow and tested on KITTI, Middlebury, and ETH3D. GGEV achieves comparable speed to RT-IGEV while offering improved generalization on unseen scenes.}
\label{fig:zero-shot}
\end{figure}

Existing real-time stereo matching methods have adopted various strategies to achieve fast inference.
These include using downsampled \cite{stereonet,bgnet} or sparse cost volume representations \cite{deeppruner,fast-acv}, lightweight aggregation networks \cite{coex}, and replacing computationally expensive 3D convolutions with 2D convolutions \cite{aanet,iinet}.
However, most existing methods rely heavily on clear and unambiguous matching cues, and struggle to effectively aggregate information in unseen domains—particularly in challenging regions such as occlusions, textureless areas, repetitive patterns, and thin structures.

Recent methods have introduced monocular foundation models (MFMs) into stereo matching, achieving remarkable zero-shot generalization performance.
FoundationStereo \cite{foundationstereo} designs a higher-capacity aggregation network to better exploit monocular priors.
MonSter \cite{monster} employs a dual-branch architecture that iteratively refines both monocular and stereo disparity estimates.
These approaches typically rely on costly backbones to extract rich and detailed features for cost volume construction, and employ complex iterative mechanisms to address the scale-shift issue between monocular and stereo.
Although these methods can improve generalization, they often overlook the critical importance of inference latency in real-world applications.
To this end, a motivating question arises: \textit{how to design a real-time stereo matching network that achieves strong generalization while maintaining high accuracy?}

To answer this question, we analyze the limitations of current geometry encoding volumes and identify two key limitations: 1) the critical regions vary significantly across disparity hypotheses; 2) the matching relationships within these regions are highly fragile due to unseen textures, occlusions, repetitive patterns, and thin structures (see Fig.~\ref{fig:attnmap}).
In this paper, we propose Generalized Geometry Encoding Volume (\textbf{GGEV}), a real-time stereo matching network that efficiently incorporates MFMs into the cost aggregation process to enhance cost volume representations.
Specifically, the proposed GGEV first constructs depth-aware features by integrating texture and depth features extracted from Depth Anything V2 \cite{depthanythingv2} using a lightweight fusion network, thereby obtaining reliable structural priors that help stabilize fragile matching relationships.
In contrast to conventional hourglass-based aggregation networks \cite{acvnet,coex} that process all disparity hypotheses uniformly, our method adaptively integrates depth structural priors into the corresponding disparity hypotheses, thereby enhancing the structural representation and generalization of the cost volume.
In particular, we first compute an affinity matrix between each disparity hypothesis and the depth feature map, where the disparity hypothesis provides positional cues and the depth features offer rich structural context.
These affinity matrices are then used to generate dynamic convolutional kernels that adaptively filter the concatenated disparity hypotheses and depth features.
Furthermore, we incorporate a combination of large and small convolution kernels to capture complementary low- and high-frequency information.

Our proposed GGEV outperforms all existing real-time stereo matching methods in both in-domain accuracy and zero-shot generalization capability.
It achieves state-of-the-art results on the KITTI 2012, KITTI 2015, and ETH3D benchmarks.
Remarkably, even when trained solely on the synthetic Scene Flow dataset, GGEV demonstrates strong cross-domain generalization to real-world scenarios, as illustrated in Fig. \ref{fig:zero-shot}.

In summary, our main contributions are:
\begin{itemize}
\item We propose a novel generalized geometry encoding volume that efficiently integrates depth priors in a lightweight manner to enhance generalization.
\item We propose a Depth-aware Dynamic Cost Aggregation (DDCA) module that adaptively generates dynamic convolution kernels based on the affinity between disparity hypotheses and depth features.
\item Our method demonstrates strong generalization to real-world scenarios, even when trained solely on the synthetic datasets.
\item Our method outperforms existing real-time approaches on public benchmarks such as KITTI 2012, KITTI 2015 and ETH3D.
\end{itemize}

\section{Related Work}

\subsection{Real-time Stereo Matching}
Recent works have focused on designing lightweight stereo matching networks while maintaining competitive accuracy.
Some methods \cite{cascade,hdrflow,chang2020attention,fadnet} attempt to construct and aggregate cost volumes at lower resolutions to reduce computational overhead.
However, this often leads to a significant drop in accuracy.
To preserve accuracy, some methods \cite{coex,arunet} still construct high-resolution cost volumes, while using context-aware activations to enable lightweight yet effective aggregation.
Alternatively, AANet \cite{aanet} replaces computationally expensive 3D convolutions with deformable 2D convolutions.
However, due to the lack of accurate structural guidance and the limited receptive field, these methods often suffer from insufficient accuracy.
Besides aggregation-based approaches, RT-IGEV \cite{igev++} achieves a balance between accuracy and efficiency by leveraging the iterative framework with a reduced number of iterations.
However, the issue of fragile matching relationships in the cost volume under unseen and challenging scenarios remains unresolved.

\subsection{Zero-shot Generalized Stereo Matching}
Zero-shot generalization in stereo matching has gained increasing attention.
Some methods \cite{dsmnet, fc} aim to learn domain-invariant representations by employing domain normalization or specialized losses.
Others mothods \cite{graftnet,chuah2022itsa} enhance generalization by avoiding shortcut learning.
In contrast to these works, RAFT-Stereo \cite{raft-stereo} enhances generalization by innovating the network framework.
Recent methods \cite{all,stereoanywhere} have further improved upon this foundation by incorporating monocular foundation models into stereo matching frameworks.
FoundationStereo \cite{foundationstereo} constructs a large-scale training dataset to support the training of larger models.
MonSter \cite{monster,monster++} employs a dual-branch architecture that iteratively refines monocular and stereo disparity estimates.
DEFOM-Stereo \cite{defom} initializes the disparity map with depth predicted by MFMs and introduces a scale update module.
However, these methods typically rely on ViT-L backbones to extract fine-grained features for cost volume construction and require complex operations to address the scale-shift inherent in monocular affine depth.
In contrast, our work emphasizes leveraging depth features to efficiently guide cost aggregation, avoiding the scale-shift issues.

\subsection{Monocular Depth Foundation Model}
MFMs~\cite{xu2025pixel,lin2025prompting,depthanythingv2} have demonstrated strong zero-shot generalization across diverse tasks without requiring target-domain fine-tuning.
MiDaS \cite{midas} pioneers zero-shot relative depth estimation, while the DINO series \cite{dino,dinov2} leverages vision transformers for dense self-supervised representation learning.
The Depth Anything series \cite{depthanythingv1,depthanythingv2} further pushes the boundary of generalization in depth estimation.
In particular, Depth Anything V2 distills student models from a large-scale teacher model using unlabeled real-world data, yielding accurate and efficient depth predictors.
This student model offers a lightweight and generalizable backbone, making it a suitable component for our framework.

\begin{figure*}[t]
\centering
\includegraphics[width=0.95\textwidth]{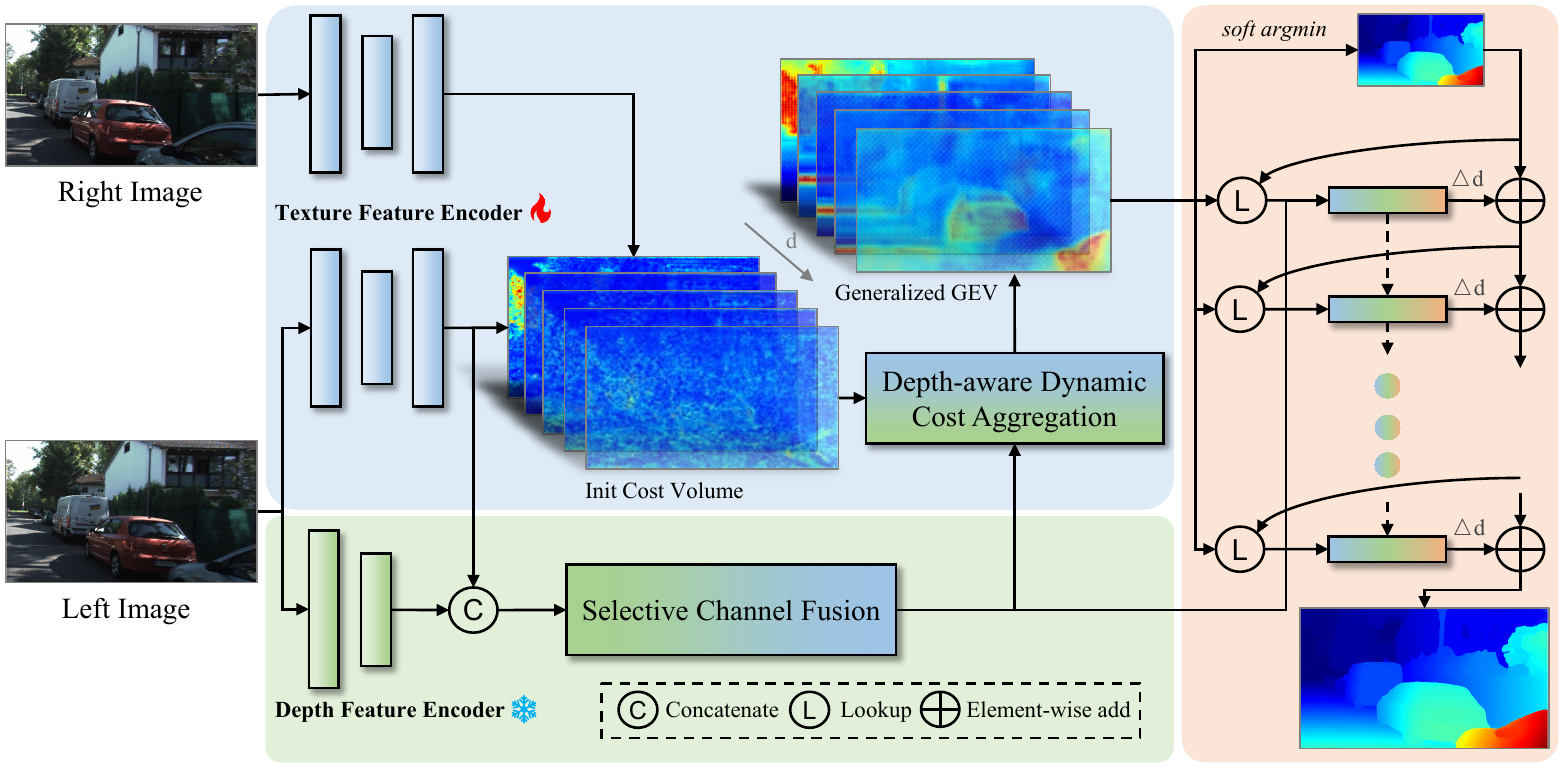} 
\caption{Overview of our proposed GGEV. The Selective Channel Fusion (SCF) module integrates texture features with depth features as a guidance for cost aggregation. Then, the Depth-aware Dynamic Cost Aggregation (DDCA) module adaptively incorporates depth structural priors to enhance the fragile matching relationships in the initial cost volume, resulting in a generalized geometry encoding volume.}
\label{fig:method}
\end{figure*}
\section{Method}

\subsection{Overall Framework}
As illustrated in the Fig. \ref{fig:method}, our network architecture consists of four main stages: multi-cue feature extraction, cost volume construction, depth-aware dynamic cost aggregation, and depth-aware iterative refinement.

\subsection{Multi-cue Feature Extraction}
The feature extraction stage consists of two complementary cues:
(a) texture features are extracted from both the left and right images to construct the cost volume;
(b) depth features are extracted from the left image to guide cost aggregation and iterative refinement.

\subsubsection{Texture Feature Encoder.}
Given rectified left and right images $\mathbf{I}_{l},\mathbf{I}_{r} \in \mathbb{R}^{{3}\times{H}\times{W}}$, we employ the MobileNetV2 pretrained on ImageNet \cite{imagenet} to extract multi-scale texture features $\mathbf{f}_{l, i},\mathbf{f}_{r, i} \in \mathbb{R}^{{C}_i \times \frac{H}{i} \times \frac{W}{i}}$,$i \in \{4,8,16\}$.

\subsubsection{Depth Feature Encoder.}
To incorporate the generalization ability of MFMs into our framework, we employ a frozen Depth Anything V2 Small to extract multi-scale depth features $\mathbf{f}_{d, i} \in \mathbb{R}^{{C}_i \times \frac{H}{i} \times \frac{W}{i}}$, $i \in \{2, 4, 8, 16\}$, using only the left image.
These features serve as depth structural priors to guide cost aggregation.
To enrich the feature representation, we additionally design a fusion module to integrate texture and depth features.

\subsubsection{Selective Channel Fusion.}
We adopt a lightweight $1 \times 1$ convolution to enable selective feature integration while preserving structural details and avoiding spatial blurring.
The Selective Channel Fusion (SCF) module takes the concatenated $\mathbf{f}_{l}$ and $\mathbf{f}_{d}$ as input and generates depth-aware prior features $\mathbf{f}_{da, i} \in \mathbb{R}^{{C}_i \times \frac{H}{i} \times \frac{W}{i}}$,$i \in \{4,8,16\}$.

\begin{figure}[t]
\centering
\includegraphics[width=0.98\columnwidth]{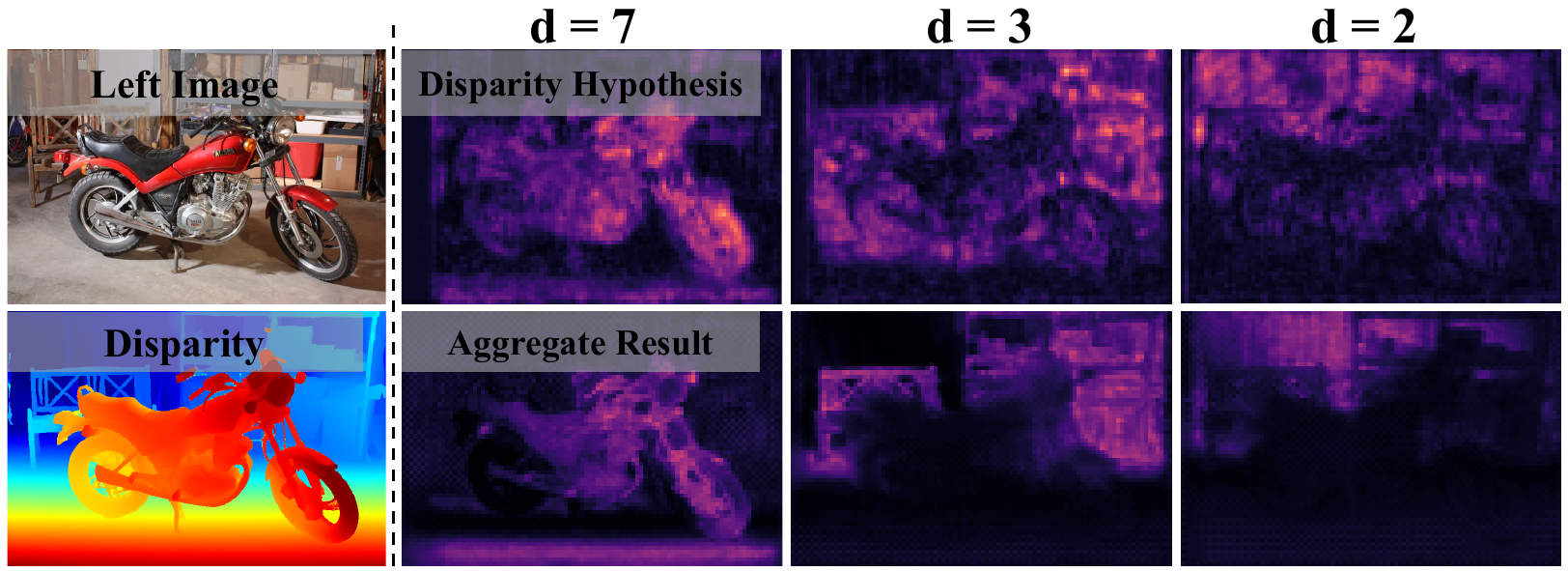} 
\caption{Effectiveness of our DDCA in generalization evaluation.
The first row show the initial cost volume features across different disparity hypotheses, which are fragile in unseen scenes and contain many mismatches.
In contrast, the second row shows the results after applying our DDCA, which effectively filters out incorrect matches and preserves accurate matching features at their corresponding disparity planes, leading to clearer and more reliable structures.
}
\label{fig:attnmap}
\end{figure}

\begin{figure}[t]
\centering
\includegraphics[width=0.95\columnwidth]{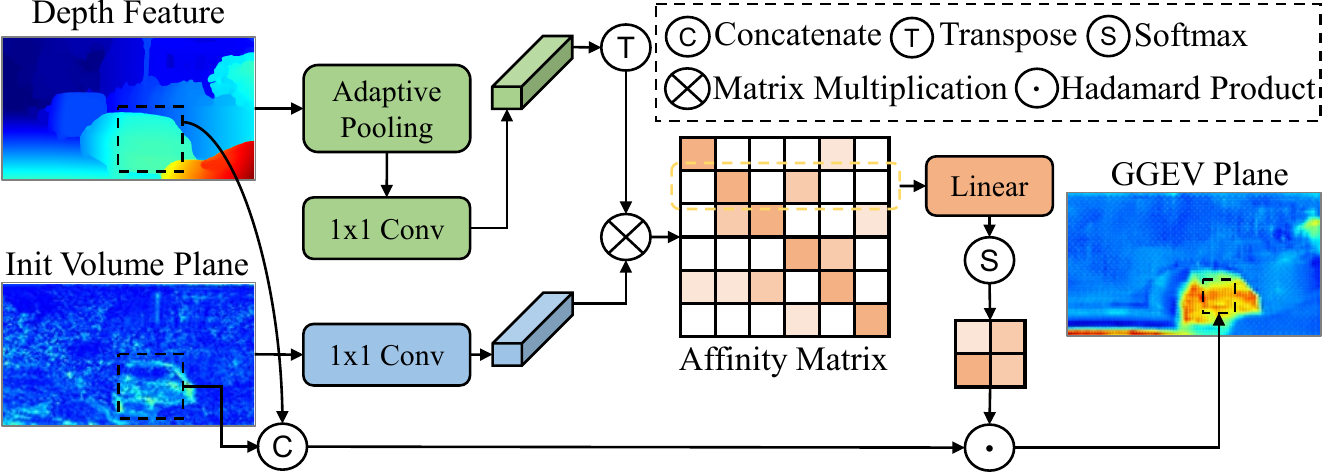} 
\caption{The architecture of proposed DDCA.}
\label{fig:DDCA}
\end{figure}

\subsection{Cost Volume Construction}
Given the texture features at 1/4 resolution $\mathbf{f}_{l,4}$ and $\mathbf{f}_{r,4}$, we construct the group-wise correlation volume $\mathbf{C}$.
\begin{equation}
\mathbf{C}\left(g, d, x, y\right)=\frac{1}{N_c / N_g}\left\langle\mathbf{f}_{l, 4}^g(x, y), \mathbf{f}_{r, 4}^g\left(x-d, y\right)\right\rangle,
\end{equation}
where $\langle \cdot, \cdot \rangle$ denotes the inner product, $d \in \mathcal{D} = \{0, 1, 2, \ldots, D/4 - 1\}$ is the disparity index, $N_c$ denotes the number of feature channels, and $N_g = 8$ is the total number of groups.

\subsection{Depth-aware Dynamic Cost Aggregation}
Along the disparity dimension, the critical regions corresponding to different disparity hypotheses often vary, and tend to be highly fragile in areas with unseen textures, occlusions, repetitive patterns, and thin structures (see Fig.~\ref{fig:attnmap}).
Treating these regions uniformly often leads to mismatches, edge blurring, loss of fine details (see Fig. \ref{fig:ablation}).
To address this, inspired by OverLoCK \cite{overlock}, we propose a novel Depth-aware Dynamic Cost Aggregation (DDCA) module that adaptively responds to cost volumes at different disparity hypotheses.
Moreover, leveraging depth-aware features as structural guidance can effectively enhance fragile matching relationships.
To dynamically model input-dependent behaviors, we propose to capture the response relationship between disparity hypotheses and depth features by computing their affinity.
These affinity values are then used to define dynamic convolution kernels, effectively injecting depth-aware structural cues into every kernel weight across different disparity hypotheses.
We employ dynamic convolution kernels to aggregate the cost maps using a sliding-window mechanism, similar to standard 2D convolutions, which makes our module lightweight and real-time.

\subsubsection{Disparity-wise Depth Structural Representation.}
As shown in Fig. \ref{fig:DDCA}, given an input disparity hypothesis map $\mathbf{C}_{d} \in \mathbb{R}^{{G}\times{H}\times{W}}$($d$ is the disparity index) and a depth-aware feature map $\mathbf{f}_{da} \in \mathbb{R}^{{C}\times{H}\times{W}}$, We first transform the input feature into two components, namely $\mathbf{Q} \in \mathbb{R}^{C \times HW}$ and $\mathbf{K} \in \mathbb{R}^{C \times S^2}$, $\mathbf{K}$ represents the aggregation of $\mathbf{f}_{da}$ into $S \times S$ regional centers via adaptive average pooling.
Simple matrix multiplications between each pair of $\mathbf{Q}$ and $\mathbf{K}$ produce the affinity matrices $\mathbf{A} \in \mathbb{R}^{HW \times S^2}$.
The process can be written as:
\begin{align}
\mathbf{Q} &= \mathrm{Re}(W_q \mathbf{C}_{d}),    \\
\mathbf{K} &= \mathrm{Re}(W_k \mathrm{Pool}(\mathbf{f}_{da})), \\
\mathbf{A} &= \mathbf{Q}^\mathrm{T} \mathbf{K},
\end{align}
where $W_q$ and $W_k$ denote $1 \times 1$ convolutional layers, and $\mathrm{Re}(\cdot)$ refers to the reshape operation.
Analogous to multi-head attention \cite{multiheadattn}, we divide $\mathbf{Q}$ and $\mathbf{K}$ along the channel dimension into $G$ groups to obtain $\mathbf{Q}^g \in \mathbb{R}^{\frac{C}{G}\times{H}{W}}$ and $\mathbf{K}^g \in \mathbb{R}^{\frac{C}{G}\times S^2}$, yielding $\mathbf{A}^g$ affinity matrices to capture the relationships within each channel group.

\subsubsection{Disparity-wise Adaptive Cost Aggregation.}
We leverage $G$ groups affinity matrices $\mathbf{A}^g$ to generate $G$ distinct $K \times K$ dynamic convolution kernels, enabling the network to adaptively aggregate features with diverse structural patterns.
First, we use another learnable linear layer ${W}_m$ to map $\mathbf{A}^g$ into corresponding convolution kernel weights $\mathbf{M}^g \in \mathbb{R}^{H W \times K^2}$, which are then normalized via a softmax operation.
Then, each row of $\mathbf{M}^g$ is reshaped into a $K \times K$ kernel, resulting in spatially adaptive convolution filters tailored to each pixel based on the input.
During the convolution operation, the channels of the fused features $\mathbf{C}_{d}$ and $\mathbf{f}_{da}$ are also divided into $G$ groups, where channels within the same group share the same dynamic kernel to reduce computational overhead.
The process can be written as:
\begin{align}
\mathbf{M}^g &= \operatorname{softmax}\left(\mathbf{A}^g {W}_m\right), \\
\mathbf{C}_{d}^{\prime} &= {\mathbf{C}_{d}} * \mathbf{M}^g_{\text {dynamic }}({\mathbf{C}_{d}},\mathbf{f}_{da}),
\end{align}
where $*$ denotes the group-wise convolution operation.
Meanwhile, we adopt a combination of large and small convolutional kernels to facilitate the fusion of low- and high-frequency information.
We rearrange the independently aggregated disparity hypotheses to reconstruct the generalized geometry encoding volume $\mathbf{C}^{\prime}$.

\subsubsection{Initial Disparity Prediction.}
We apply the \textit{soft-argmin} to $\mathbf{C}^{\prime}$ to regress the initial disparity estimation.
\begin{equation}
\mathbf{d}_0 = \sum_{d\in\mathcal{D}}d\times Softmax(\mathbf{C}^{\prime}(d)),
\end{equation}
where $\mathbf{d}_0$ is at 1/4 scale of the original image resolution.

\subsection{Depth-aware Iterative Refinement}
Given $\mathbf{d}_0$, we employ an iterative GRU to progressively refine the disparity map.
The hidden state ${h}_0$ is initialized using the depth feature $\mathbf{f}_{da, 4}$, injecting structural priors into the recurrent refinement process.
The single-layer GRU updates its hidden state based on the latest disparity $\mathbf{d}_{k}$ and the geometry features $\mathbf{f}_{G}$ indexed from $\mathbf{C}^{\prime}$.
The update operator can be formulated as:
\begin{align}
z_k&=\sigma(\mathrm{Conv}([h_{k-1},x_k],W_z)),   \\
r_k&=\sigma(\mathrm{Conv}([h_{k-1},x_k],W_r)),    \\
\tilde{h}_k&=\tanh(\mathrm{Conv}([r_k\odot h_{k-1},x_k],W_h)),   \\
h_k&=(1-z_k)\odot h_{k-1}+z_k\odot\tilde{h}_k,
\end{align}
where $x_k$ denotes the concatenation of $\mathbf{d}_{k}$ and $\mathbf{f}_{G}$; $\odot$ denotes element-wise product; $\sigma$ denotes sigmoid; $W_z$, $W_r$ and $W_h$ are the parameters of the network.
Based on the ${h}_k$, we decode a residual disparity $\Delta \mathbf{d}_k$ using two convolutional layers, and update the current disparity accordingly.
\begin{equation}
\mathbf{d}_{k+1}=\mathbf{d}_{k}+\Delta \mathbf{d}_k.
\end{equation}

\subsubsection{Spatial Upsampling.}
We leverage depth features to assist in recovering full-resolution disparity map.
We first convolve the ${h}_k$ to generate features and upsample them to half resolution.
The upsampled features are then concatenated with the depth features $\mathbf{f}_{d}$ from left image to produce a weight map $W \in \mathbb{R}^{{H}\times{W}\times{9}}$.
Finally, we obtain the full-resolution disparity map by performing a weighted combination over the local neighboring points of the low-resolution disparity map $\mathbf{d}_k$.

\subsection{Loss Function}
We use Smooth L1 loss to supervise the initial disparity $\mathbf{d}_0$ regression from GGEV, and employ L1 loss to supervise the disparity $
\left\{\mathbf{d}_i\right\}_{i=1}^N$ estimated through iterative refinement.
\begin{equation}
\mathcal{L}=\left|\mathbf{d}_0-\mathbf{d}_{gt}\right|_{smooth}+\sum_{i=1}^N \gamma^{N-i}\left\|\mathbf{d}_i-\mathbf{d}_{gt}\right\|_1,
\end{equation}
where $\mathbf{d}_{gt}$ denotes the ground truth disparity, $\gamma = 0.9$ is a decay factor, and $N$ is the number of iterations used during training.

\section{Experiments}

\subsection{Implementation Details}
We implement our GGEV with PyTorch and perform our experiments using NVIDIA RTX 3090 GPUs.
For all training, we use the AdamW optimizer and clip gradients to the range [-1, 1], and adopt a one-cycle learning rate schedule.
We perform 11 update iterations during training and use 8 iterations during inference.
Following standard \cite{psmnet,s2m2,diving}, we pretrain GGEV on the Scene Flow \cite{dispnetc} for most experiments.
For finetuning on KITTI \cite{kitti2012,kitti2015}, we use the mixed dataset of KITTI 2012 and KITTI 2015.
For finetuning on ETH3D \cite{eth3d}, we follow the CREStereo \cite{crestereo} and GMStereo \cite{unistereo}, using a collection of public stereo datasets.
Detailed information is provided in the Supplemental Material.

\begin{table}[t]
\small
\centering
{\setlength{\tabcolsep}{0.5mm}
\begin{tabular}{@{}cllcccc@{}}
\toprule
\multicolumn{2}{l}{\textbf{Training Set}} & \multicolumn{5}{c}{\textbf{Scene Flow}} \\ \midrule
\multicolumn{1}{l|}{\multirow{2}{*}{Target}} & \multicolumn{2}{l|}{\multirow{2}{*}{Method}} & \multicolumn{2}{c|}{KITTI} & \multicolumn{1}{c|}{Middlebury} & \multirow{2}{*}{ETH3D} \\
\multicolumn{1}{l|}{} & \multicolumn{2}{l|}{} & 2012 & \multicolumn{1}{c|}{2015} & \multicolumn{1}{c|}{quarter} &  \\ \midrule
\multicolumn{1}{c|}{\multirow{5}{*}{\rotatebox{90}{\textit{Accuracy}}}} & 
\multicolumn{2}{l|}{RAFT-Stereo \citeyearpar{raft-stereo}} & 4.5 & \multicolumn{1}{c|}{5.7} & \multicolumn{1}{c|}{9.3} & 3.2 \\
\multicolumn{1}{c|}{} & \multicolumn{2}{l|}{FC-GANet \citeyearpar{fc}} & 4.6 & \multicolumn{1}{c|}{5.3} & \multicolumn{1}{c|}{7.8} & 5.8 \\
\multicolumn{1}{c|}{} & \multicolumn{2}{l|}{DEFOM-Stereo \citeyearpar{defom}} & 3.7 & \multicolumn{1}{c|}{4.9} & \multicolumn{1}{c|}{5.6} & 2.3 \\
\multicolumn{1}{c|}{} & \multicolumn{2}{l|}{DEFOM-Stereo (ViT-S)} & 4.2 & \multicolumn{1}{c|}{5.3} & \multicolumn{1}{c|}{6.3} & 2.6 \\
\multicolumn{1}{c|}{} & \multicolumn{2}{l|}{FoundationStereo \citeyearpar{foundationstereo}} & 3.2 & \multicolumn{1}{c|}{4.9} & \multicolumn{1}{c|}{-} & 1.8 \\ \midrule
\multicolumn{1}{c|}{\multirow{7}{*}{\rotatebox{90}{\textit{Speed}}}} & 
\multicolumn{2}{l|}{DeepPrunerFast \citeyearpar{deeppruner}} & 16.8 & \multicolumn{1}{c|}{15.9} & \multicolumn{1}{c|}{18.3} & 11.0 \\
\multicolumn{1}{c|}{} & \multicolumn{2}{l|}{CoEx \citeyearpar{coex}} & 13.5 & \multicolumn{1}{c|}{10.6} & \multicolumn{1}{c|}{14.5} & 9.0 \\
\multicolumn{1}{c|}{} & \multicolumn{2}{l|}{BGNet+ \citeyearpar{bgnet}} & 5.3 & \multicolumn{1}{c|}{6.6} & \multicolumn{1}{c|}{11.2} & 10.3 \\
\multicolumn{1}{c|}{} & \multicolumn{2}{l|}{Fast-ACVNet \citeyearpar{fast-acv}} & 12.4 & \multicolumn{1}{c|}{10.6} & \multicolumn{1}{c|}{13.5} & 7.9 \\

\multicolumn{1}{c|}{} & \multicolumn{2}{l|}{IINet \citeyearpar{iinet}} & 11.6 & \multicolumn{1}{c|}{8.5} & \multicolumn{1}{c|}{-} & - \\
\multicolumn{1}{c|}{} & \multicolumn{2}{l|}{RT-IGEV \citeyearpar{igev++}} & 5.8 & \multicolumn{1}{c|}{6.6} & \multicolumn{1}{c|}{7.8} & 5.8 \\
\multicolumn{1}{c|}{} & \multicolumn{2}{l|}{GGEV (Ours)} & \textbf{4.1} & \multicolumn{1}{c|}{\textbf{5.5}} & \multicolumn{1}{c|}{\textbf{6.5}} & \textbf{2.8} \\ \midrule
\multicolumn{2}{l}{\textbf{Training Set}} & \multicolumn{5}{c}{\textbf{Scene Flow + CREStereo + Tartan Air}} \\ \midrule
\multicolumn{1}{l|}{\multirow{2}{*}{}} & \multicolumn{2}{l|}{RT-IGEV \citeyearpar{igev++}} & 4.0 & \multicolumn{1}{c|}{5.4} & \multicolumn{1}{c|}{8.6} & 3.4 \\
\multicolumn{1}{l|}{} & \multicolumn{2}{l|}{GGEV (Ours)} & \textbf{3.6} & \multicolumn{1}{c|}{\textbf{4.7}} & \multicolumn{1}{c|}{\textbf{5.7}} & \textbf{2.2} \\ \bottomrule
\end{tabular}
}
\caption{Zero-shot generalization from synthetic to real. Standard threshold-based error metrics are employed for evaluation: 3-pixel for KITTI, 2-pixel for Middlebury, and 1-pixel for ETH3D. \textbf{Bold}: Best.}
\label{tab:zero-shot}
\end{table}

\begin{table*}[t]
\centering
\begin{tabular}{@{}c|l|cccccc|ccc|c@{}}
\toprule
\multirow{3}{*}{Target} & \multirow{3}{*}{Method} & \multicolumn{6}{c|}{KITTI 2012} & \multicolumn{3}{c|}{KITTI 2015} & \multirow{3}{*}{\begin{tabular}[c]{@{}c@{}}Time\\ (ms)\end{tabular}} \\ \cmidrule(lr){3-11}
 &  & \multicolumn{2}{c}{2-pixels} & \multicolumn{2}{c}{3-pixels} & \multicolumn{2}{c|}{EPE} & D1-bg & D1-fg & D1-all &  \\ \cmidrule(lr){3-11}
 &  & noc & all & noc & all & noc & all & \multicolumn{3}{c|}{all} &  \\ \midrule
\multirow{4}{*}{\rotatebox{90}{\textit{Accuracy}}}
 & RAFT-Stereo \citeyearpar{raft-stereo} & 1.92 & 2.42 & 1.30 & 1.66 & 0.4 & 0.5 & 1.58 & 3.05 & 1.82 & 380 \\
 & IGEV \citeyearpar{igev} & 1.71 & 2.17 & 1.12 & 1.44 & 0.4 & 0.4 & 1.38 & 2.67 & 1.59 & 180 \\
 & Selective-IGEV \citeyearpar{selective} & 1.59 & 2.05 & 1.07 & 1.38 & 0.4 & 0.4 & 1.33 & 2.61 & 1.55 & 240 \\
 & Moster \citeyearpar{monster} & 1.36 & 1.75 & 0.84 & 1.09 & 0.4 & 0.4 & 1.13 & 2.81 & 1.41 & 450 \\ \midrule
\multirow{12}{*}{\rotatebox{90}{\textit{Speed}}}
 & DeepPrunerFast \citeyearpar{deeppruner} & - & - & - & - & - & - & 2.32 & 3.91 & 2.59 & 50 \\
 & AANet \citeyearpar{aanet} & 2.30 & 2.96 & 1.55 & 2.05 & 0.4 & 0.5 & 1.65 & 3.96 & 2.03 & 60 \\
 & HITNet \citeyearpar{hitnet} & 2.00 & 2.65 & 1.41 & 1.89 & 0.4 & 0.5 & 1.74 & {3.20} & 1.98 & 20 \\
 & BGNet+ \citeyearpar{bgnet} & 2.78 & 3.35 & 1.62 & 2.03 & 0.5 & 0.6 & 1.81 & 4.09 & 2.19 & 35 \\
 & DecNet \citeyearpar{decomposition} &  & - & - & - & - & - & 2.07 & 3.87 & 2.37 & 50 \\
 & CoEx \citeyearpar{coex} & 2.54 & 3.09 & 1.55 & 1.93 & 0.5 & 0.5 & 1.74 & 3.41 & 2.02 & 27 \\
 & TemporalStereo \citeyearpar{temporalstereo} & - & - & - & - & - & - & 1.61 & \textbf{2.78} & 1.81 & 45 \\
 & Fast-ACVNet+ \citeyearpar{fast-acv} & 2.39 & 2.97 & 1.45 & 1.85 & 0.5 & 0.5 & 1.70 & 3.53 & 2.01 & 45 \\
 & IINet \citeyearpar{iinet} & 2.76 & 3.34 & 1.81 & 2.21 & 0.5 & 0.5 & 2.02 & 3.39 & 2.25 & 26 \\		
 & RT-IGEV \citeyearpar{igev++} & {1.93} & {2.51} & 1.29 & 1.68 & 0.4 & 0.5 & {1.48} & 3.37 & 1.79 & 40 \\ 
 & BANet-3D \citeyearpar{banet} & 2.08 & 2.71 & {1.27} & 1.72 & 0.5 & 0.5 & 1.52 & 3.02 & {1.77} & 30 \\
 & GGEV (Ours) & \textbf{1.66} & \textbf{2.17} & \textbf{1.10} & \textbf{1.44} & 0.4 & 0.4 & \textbf{1.38} & 3.28 & \textbf{1.70} & 47 \\ \bottomrule
\end{tabular}
\caption{Quantitative evaluation on KITTI 2012 and KITTI 2015. \textbf{Bold}: Best.}
\label{tab:KITTI-benchmark}
\end{table*}

\subsection{Zero-Shot Generalization}
We follow MonSter by training our model on the Scene Flow and directly evaluating it on four real-world datasets.
The results are shown in Tab. \ref{tab:zero-shot}, our GGEV achieves the best performance among all real-time models and demonstrates generalization capability comparable to that of some accuracy-oriented methods.
This improvement is attributed to the incorporation of monocular depth priors and the adaptive handling of different disparity hypotheses.
Specifically, compared to the SOTA real-time method RT-IGEV, our GGEV reduces the error rates by 29\% on KITTI 2012, 16\% on KITTI 2015, and 16\% on Middlebury-quarter. More notably, it achieves a 51\% error reduction on the real-world indoor-outdoor ETH3D dataset.
Our GGEV also outperforms domain generalization methods, such as FC-GANet.
Using the same ViT-S backbone, our GGEV achieves comparable performance to DEFOM-Stereo (ViT-S, 255ms), while reducing inference time by 81\%, highlighting the efficiency of our approach.

By adding more synthetic datasets, the generalization performance of our GGEV is further improved, still surpassing that of RT-IGEV under the same training conditions.
This demonstrates that high-quality training data can better guide our network in learning effective cost aggregation.

\begin{figure}[t]
\centering
\includegraphics[width=0.98\columnwidth]{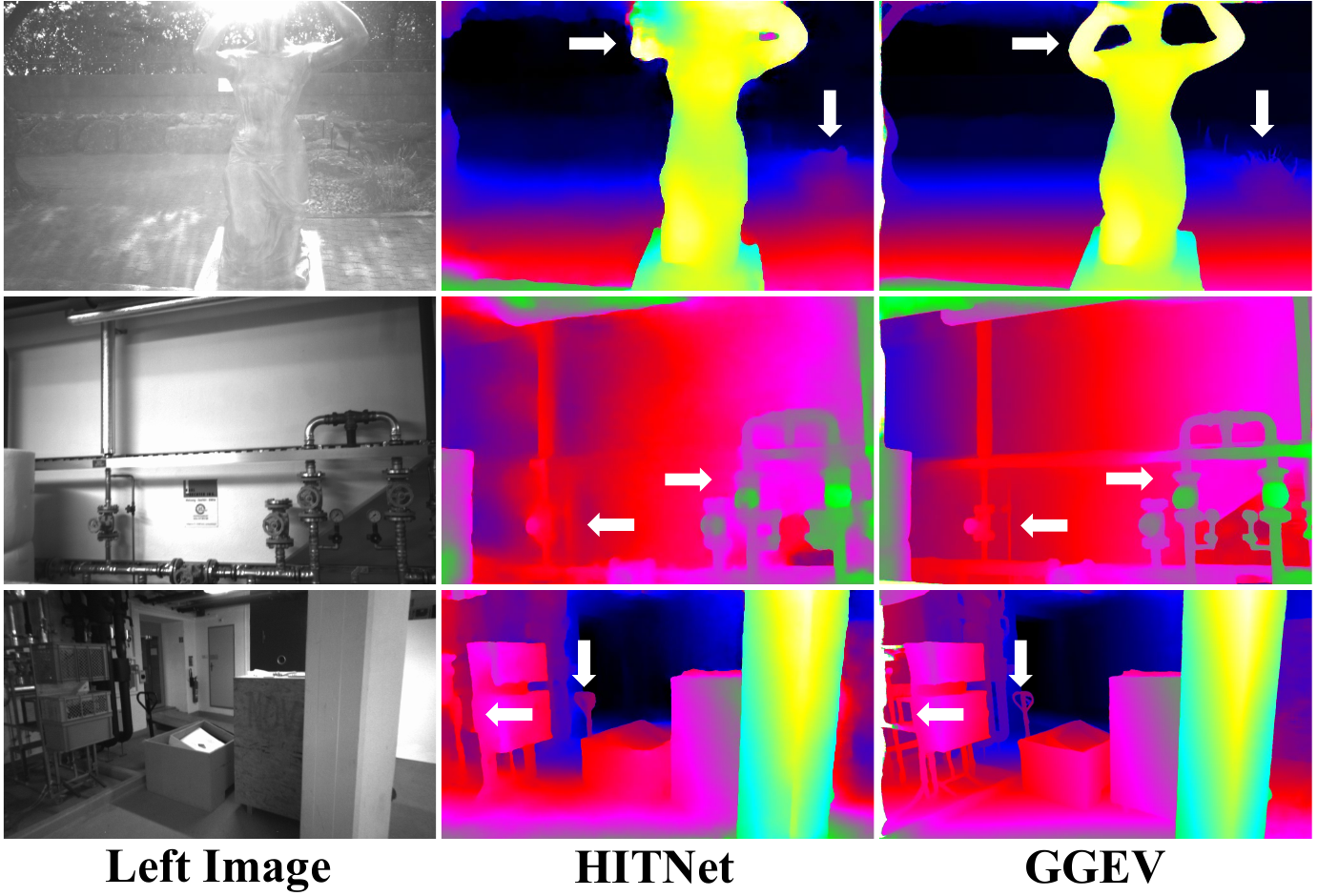} 
\caption{Qualitative comparison on ETH3D.}
\label{fig:eth3d-benchmark}
\end{figure}

\subsection{Benchmark Comparisons}
To demonstrate the outstanding performance of our method, we conduct comprehensive comparisons with prior methods on three widely-used stereo benchmarks: KITTI 2012, KITTI 2015, and ETH3D.
The fine-tuning settings are in the Supplementary Material.

\subsubsection{KITTI.}
As shown in Tab. \ref{tab:KITTI-benchmark} and Fig. \ref{fig:kitti-benchmark}, among all real-time models, our method achieves $1^{st}$ performance.
On the KITTI 2012, our proposed GGEV outperforms both RT-IGEV and BANet-3D by 13\% on the 2-noc and 3-noc metrics.
On the KITTI 2015, our proposed GGEV achieves the top performance on both D1-bg and D1-all metrics.

\subsubsection{ETH3D.}
As shown in Tab. \ref{tab:ETH3D-benchmark} and Fig. \ref{fig:eth3d-benchmark}, our GGEV significantly outperforms all existing real-time stereo matching methods across all evaluation metrics, with nearly 50\% error reduction on each metric.
Notably, our method surpasses both GMStereo and Selective-IGEV on the Bad 1.0 metric, while requiring less than one-fourth of their inference time.

\begin{table}[t]
\small
\centering
{\setlength{\tabcolsep}{0.5mm}
\begin{tabular}{@{}c|l|cccc@{}}
\toprule
\multicolumn{1}{l|}{\multirow{2}{*}{Tgt}} & \multirow{2}{*}{Method} & \multicolumn{4}{c}{ETH3D} \\
\multicolumn{1}{l|}{} &  & Bad 0.5 & Bad 1.0 & Bad 2.0 & AvgErr \\ \midrule
\multirow{4}{*}{\rotatebox{90}{\textit{Accuracy}}}
& RAFT-Stereo \citeyearpar{raft-stereo} & 7.04 & 2.44 & 0.44 & 0.18 \\
 & GMStereo \citeyearpar{unistereo} & 5.94 & 1.83 & 0.25 & 0.19 \\
 & Selective-IGEV \citeyearpar{selective} & 3.06 & 1.23 & 0.22 & 0.12 \\
 & FoundationStereo \citeyearpar{foundationstereo} & 1.26 & 0.26 & 0.08 & 0.09 \\ \midrule
\multirow{3}{*}{\rotatebox{90}{\textit{Speed}}}
 & Fast-ACVNet \citeyearpar{fast-acv} & 14.25 & 5.62 & 1.41 & 0.31 \\
 & HITNet \citeyearpar{hitnet} & {7.83} & {2.79} & {0.80} & {0.20} \\
 & GGEV (Ours) & \textbf{3.70} & \textbf{1.19} & \textbf{0.34} & \textbf{0.14} \\ \bottomrule
\end{tabular}
}
\caption{Quantitative evaluation on ETH3D. \textbf{Bold}: Best.}
\label{tab:ETH3D-benchmark}
\end{table}

\begin{table*}[t]
\centering
{\setlength{\tabcolsep}{1mm}
\begin{tabular}{@{}l|ccc|c|c|c|c|c|cc@{}}
\toprule
\multirow{2}{*}{Models} & \multicolumn{3}{c|}{Proposed Modules} & Scene Flow & KITTI 2012 & KITTI 2015 & Middlebury & ETH3D & \multirow{2}{*}{\begin{tabular}[c]{@{}c@{}}Params.\\ (M)\end{tabular}} & \multirow{2}{*}{\begin{tabular}[c]{@{}c@{}}Time\\ (ms)\end{tabular}} \\
 & DFE & SCF & DCA & EPE & D1 & D1 & Bad 2.0 & Bad 1.0 &  &  \\ \midrule
Baseline &  &  &  & 0.54 & 6.63 & 8.01 & 7.84 & 7.54 & 3.60 & 30 \\ \midrule
+DFE (ViT-S) & $\checkmark$ &  &  & 0.52 & 4.40 & 6.32 & 6.47 & 5.02 & 3.57 & 37 \\
+DFE+SCF & $\checkmark$ & $\checkmark$ &  & 0.49 & 5.14 & 7.58 & 5.57 & 4.85 & 3.65 & 38 \\
+DCA &  &  & $\checkmark$ & 0.47 & 6.80 & 6.75 & 7.73 & 5.61 & 3.63 & 39 \\
Full Model (ViT-L) & $\checkmark$ & $\checkmark$ & $\checkmark$ & 0.45 & 4.20 & 5.65 & 4.41 & 1.69 & 3.75 & 110 \\
Full Model (ViT-S) & $\checkmark$ & $\checkmark$ & $\checkmark$ & 0.46 & 4.11 & 5.56 & 6.53 & 2.84 & 3.68 & 47 \\ \bottomrule
\end{tabular}
}
\caption{Ablation study of proposed networks on the Scene Flow test set and zero-shot generation. The parameters counted here are the trainable ones. The run-time is measured at the KITTI resolution of 1248 $\times$ 384.}
\label{tab:ablation}
\end{table*}

\begin{table}[t]
\centering
 \begin{tabular}{@{}l|cccc@{}}
\toprule
\multirow{2}{*}{Method} & \multicolumn{4}{c}{KITTI 2012 (Reflective)} \\
 & 2-noc & 2-all & 3-noc & 3-all \\ \midrule
RAFT-Stereo \citeyearpar{raft-stereo} & 8.41 & 9.87 & 5.40 & 6.48 \\
CREStereo \citeyearpar{crestereo} & 9.71 & 11.26 & 6.27 & 7.27 \\
HITNet \citeyearpar{hitnet} & 9.75 & 11.85 & 5.91 & 7.54 \\
RT-IGEV \citeyearpar{igev++} & 9.56 & 11.54 & 5.76 & 7.26 \\ 
BANet-3D \citeyearpar{banet} & 9.64 & 11.97 & 5.37 & 7.07 \\
GGEV (Ours) & \textbf{7.33} & \textbf{9.27} & \textbf{4.04} & \textbf{5.34} \\ \bottomrule
\end{tabular}
\caption{Evaluation on reflective (Ill-Posed) regions of the KITTI 2012 benchmark. \textbf{Bold}: Best.}
\label{tab:reflective}
\end{table}

\begin{figure}[t]
\centering
\includegraphics[width=0.98\columnwidth]{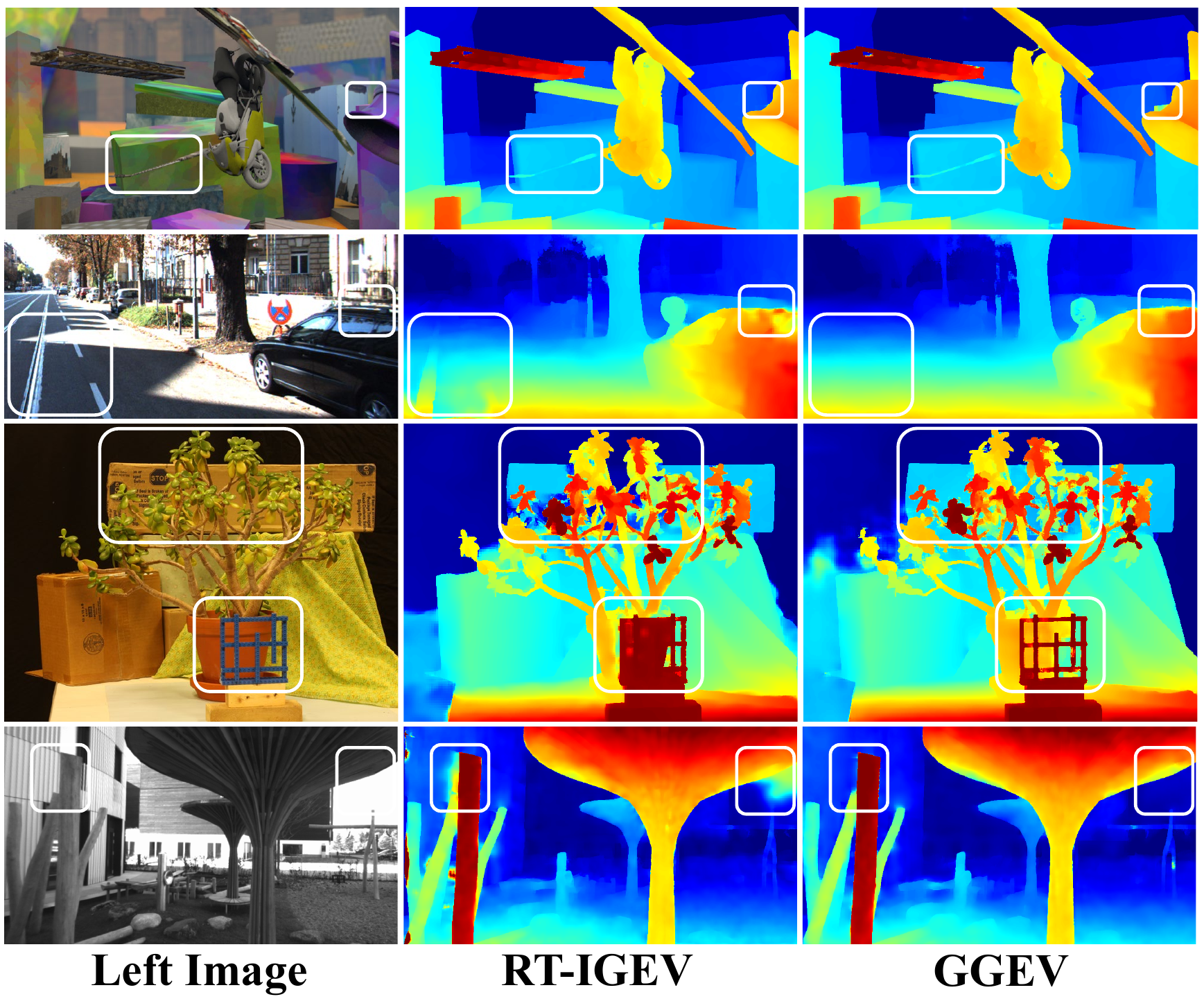} 
\caption{Qualitative comparison of disparity regression results between the GEV from RT-IGEV and our generalized GEV on Scene Flow, KITTI, Middlebury and ETH3D. GGEV shows improved performance in edges, occlusions, and textureless areas, even without iterative operations.}
\label{fig:ablation}
\end{figure}

\subsection{Ablation Study}
We conduct a series of ablation studies to verify the effectiveness and rationality of the proposed modules.
We use a simplified RT-IGEV as our baseline by reducing the aggregation depth to two downsampling stages and the number of inference iterations is set to 8.
All model variants are trained for 200k steps on the Scene Flow and evaluated under both in-domain and zero-shot generalization settings.
Additionally, we report the inference time and trainable parameters for each variant.
The results are shown in Tab. \ref{tab:ablation}.

\subsubsection{Effecitveness of Depth Feature Encoder.}
The incorporation of depth features enhances generalization performance, owing to the strong generalization of the pretrained MFMs. However, treating all disparity hypotheses uniformly fails to fully leverage the potential of the depth features.

\subsubsection{Effecitveness of Selective Channel Fusion.}
The fusion of texture and depth features enhances the model's in-domain fitting, but its effect on generalization is mixed.

\subsubsection{Effecitveness of Dynamic Cost Aggregation.}
We deploy Dynamic Cost Aggregation module guided by texture features, which significantly improves in-domain performance.
However, texture features are sensitive to textureless regions and appearance variations, resulting in limited improvements in generalization.

\subsubsection{Effecitveness of All Components.}
By integrating all the proposed modules, we observe comprehensive improvements in both accuracy and generalization.
The SCF introduces the generalization capability of MFMs into the cost aggregation, while the DDCA adaptively incorporates depth features to enhance fragile matching relationships.
Qualitative results of the initial disparity from RT-IGEV and GGEV are presented in Fig.~\ref{fig:ablation}, where our full model demonstrates both high accuracy and strong generalization capability, even without iterative refinement.
Additionally, employing a ViT-L backbone further boosts model performance, while introducing higher inference latency.

\subsubsection{Trainbale Parameters.}
Our full model (ViT-S) introduces only a marginal increase of 0.08M (+2\%) trainable parameters compared to the baseline.
Specifically, the proposed DFE and SCF add 0.05M parameters, while the DCA contributes an additional 0.03M.

\subsubsection{Inference Times.}
Our full model (ViT-S) satisfies HITNet’s real-time constraint of 100ms, with a 17ms increase over the baseline, including 8ms from the DFE and SCF and 9ms from the DCA, while achieving comparable speed to RT-IGEV and delivering significantly better performance.

\begin{figure}[t]
\centering
\includegraphics[width=0.98\columnwidth]{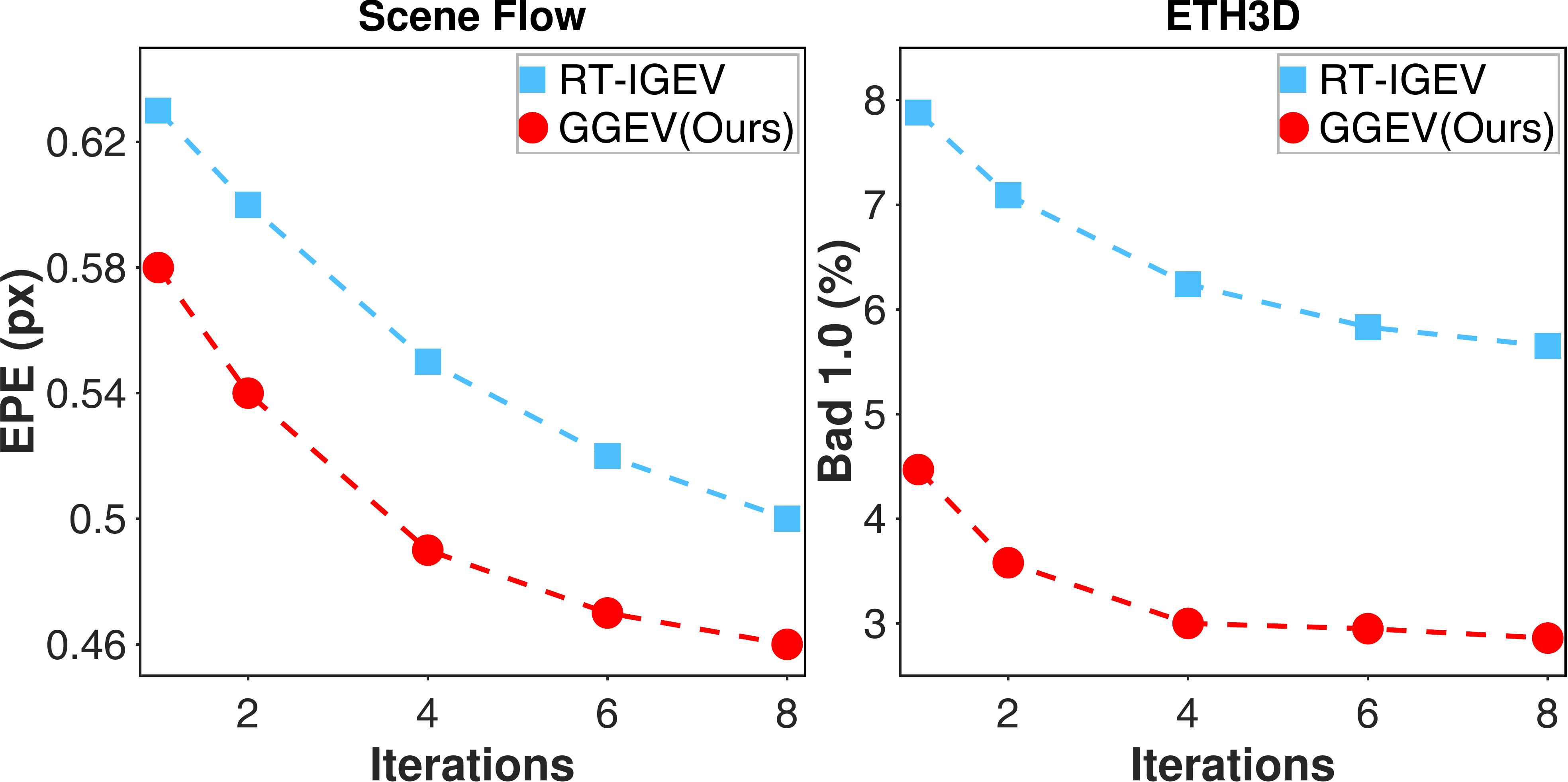} 
\caption{Performance comparison with different numbers of inference iterations.}
\label{fig:iters}
\end{figure}

\subsubsection{Number of Iterations.}
Our GGEV achieves better in-domain and zero-shot performance with fewer iterations.
As shown in Fig. \ref{fig:iters}, our method consistently outperforms RT-IGEV under the same number of iterations.

\begin{figure}[t]
\centering
\includegraphics[width=0.98\columnwidth]{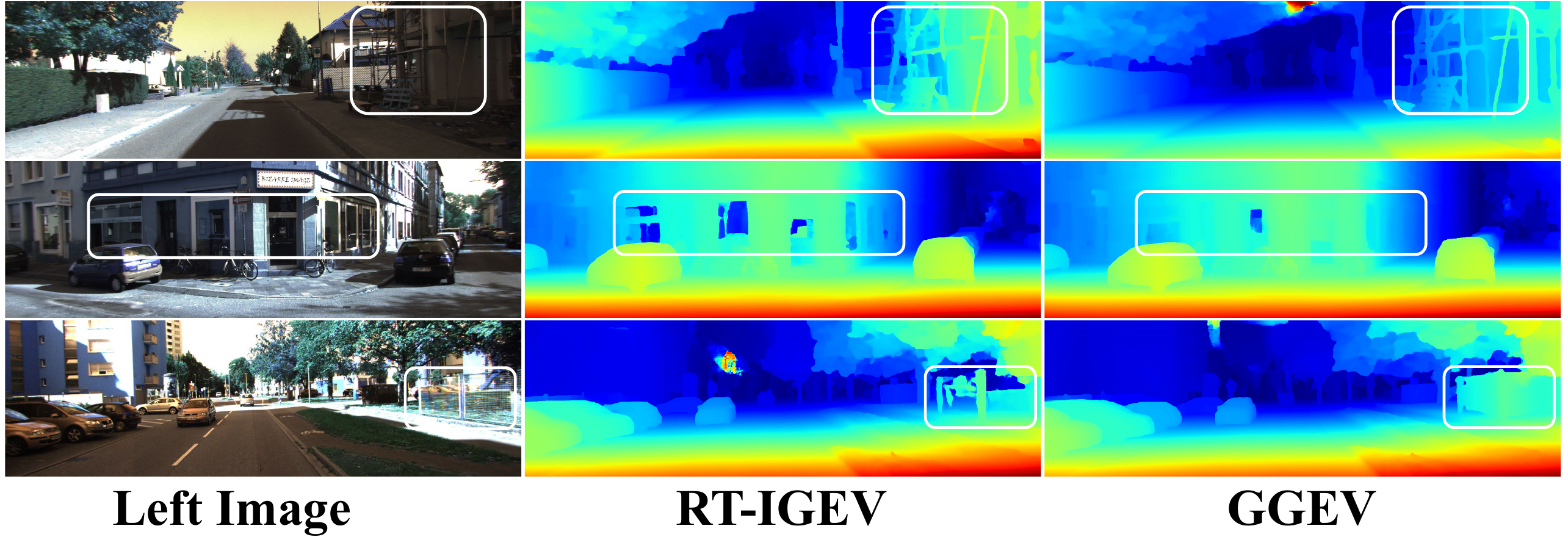} 
\caption{Qualitative comparison on KITTI. Our GGEV shows significant improvement in challenging regions such as fine structures and reflective surfaces.}
\label{fig:kitti-benchmark}
\end{figure}

\subsection{Performance in Ill-posed Regions}
We evaluate GGEV on reflective areas in KITTI 2012.
As shown in Tab. \ref{tab:reflective}, our GGEV achieves the best performance among all real-time stereo matching methods, even surpassing several accuracy-oriented approaches.
Compared to RAFT-Stereo, it achieves improvements of 12\% and 25\% on the 2-noc and 3-noc metrics, respectively.

\section{Conclusion}
We presented GGEV, a real-time stereo matching framework that achieves impressive zero-shot generalization.
The proposed Selective Channel Fusion and Depth-aware Dynamic Cost Aggregation modules extract depth-aware features as generalized structural guidance and adaptively aggregate disparity-specific critical regions, collaboratively generating a generalized geometry encoding volume.
Experimental results demonstrate that our method consistently outperforms all existing real-time approaches.
Future work could explore leveraging metric depth foundation models to provide more accurate depth guidance or extending our method to real-time video stereo matching.

\section{Acknowledgments}
This research is supported by the National Key R\&D Program of China (2024YFE0217700), the National Natural Science Foundation of China (62472184,623B2036), the Fundamental Research Funds for the Central Universities, and the Innovation Project of Optics Valley Laboratory (Grant No. OVL2025YZ005).

\bibliography{aaai2026}

\clearpage
\appendix

\clearpage
\appendix

\twocolumn[
\begin{center}
{\LARGE \bf Generalized Geometry Encoding Volume for Real-time Stereo Matching \par}  
\vspace{8pt}
{\Large Supplementary Material}               
\end{center}
\vspace{1em}
]


\begin{figure}[t]
\centering
\includegraphics[width=0.98\columnwidth]{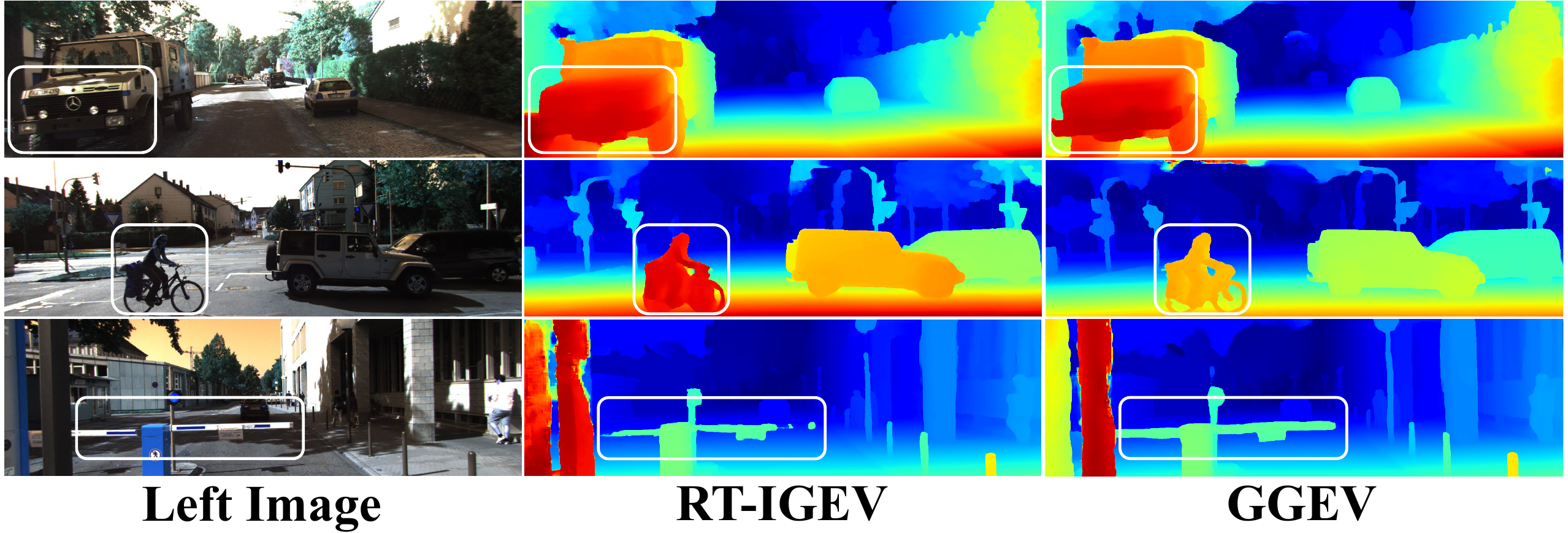} 
\caption{Qualitative comparison on KITTI.}
\label{fig:kitti-benchmark}
\end{figure}


\subsection{Datasets and Metrics}
\subsubsection{Scene Flow.}
Scene Flow \cite{dispnetc} is a synthetic dataset comprising 35,454 training pairs and 4,370 testing pairs, with dense ground-truth disparity annotations.
We use the Finalpass subset for both training and evaluation, as it better resembles real-world imagery.
The end-point error (EPE) is adopted as the evaluation metric.
During training, we randomly crop images to $320 \times 768$ with a batch size of 12, and apply data augmentation techniques including asymmetric chromatic and spatial augmentations.

\subsubsection{KITTI.}
KITTI 2012 \cite{kitti2012} and KITTI 2015 \cite{kitti2015} are real-world driving datasets.
KITTI 2012 contains 194 training pairs and 195 testing pairs, and KITTI 2015 contains 200 training pairs and 200 testing pairs.
Both datasets provide sparse ground-truth disparities obtained with LiDAR.
KITTI 2012 is evaluated using x-noc/x-all and EPE-noc/EPE-all, while KITTI 2015 reports the percentage of pixels with errors exceeding 3 pixels in background (D1-bg), foreground (D1-fg), and all regions (D1-all).
We finetuned the model pretrained on Scene Flow for 50k steps with a batch size of 8, using a mixed dataset of KITTI 2012 and KITTI 2015.
The qualitative comparison results are shown in Fig. \ref{fig:kitti-benchmark}.

\subsubsection{ETH3D.}
ETH3D \cite{eth3d} is a grayscale stereo dataset with 27 training and 20 testing pairs from both indoor and outdoor scenes.
Evaluation is based on the percentage of pixels with errors exceeding 1 pixel (Bad 1.0) and 2 pixels (Bad 2.0).
Following the training strategy of CREStereo and GMStereo, we train our model on a collection of public stereo datasets.
We first finetune the Scene Flow-pretrained model for 300k steps with a batch size of 8 and a crop size of $384 \times 512$ on a mixed dataset consisting of TartanAir, CREStereo, Scene Flow, Sintel Stereo, InStereo2k, and ETH3D.
Subsequently, we further finetune the model for another 100k steps with the same batch size on a mixed dataset comprising CREStereo, InStereo2k, and ETH3D.

\subsubsection{Middlebury.}
Middlebury V3 \cite{middlebury} is an indoor stereo dataset with 15 training and 15 testing pairs, some of which exhibit inconsistent illumination or color.
To evaluate cross-domain generalization, we use the quarter-resolution version of the dataset.
Performance is measured using the Bad 2.0 metric.

\begin{table}[t]
\centering
\begin{tabular}{@{}l|cc@{}}
\toprule
Method & EPE (px) & Time (ms) \\ \midrule
DeepPrunerFast \citeyearpar{deeppruner} & 0.97 & 61 \\
AANet \citeyearpar{aanet} & 0.87 & 62 \\
BGNet \citeyearpar{bgnet} & 1.17 & 28 \\
DecNet \citeyearpar{decomposition} & 0.84 & 50 \\
CoEx \citeyearpar{coex} & 0.69 & 33 \\
Fast-ACVNet+ \citeyearpar{fast-acv} & 0.59 & 45 \\
IINet \citeyearpar{iinet} & 0.54 & 26 \\
RT-IGEV \citeyearpar{igev++} & 0.50 & 40 \\ 
BANet-3D \citeyearpar{banet} & 0.51 & 30 \\
GGEV (Our, Iters=2) & 0.54 & 35 \\
GGEV (Our, Iters=4) & 0.49 & 39 \\
GGEV (Our, Iters=6) & 0.47 & 44 \\
GGEV (Our, Iters=8) & \textbf{0.46} & 47 \\ \bottomrule
\end{tabular}
\caption{Comparisons with other real-time methods on the Scene Flow test set. \textbf{Bold}: Best.}
\label{tab:sceneflow}
\end{table}

\begin{figure*}[t]
\centering
\includegraphics[width=0.95\textwidth]{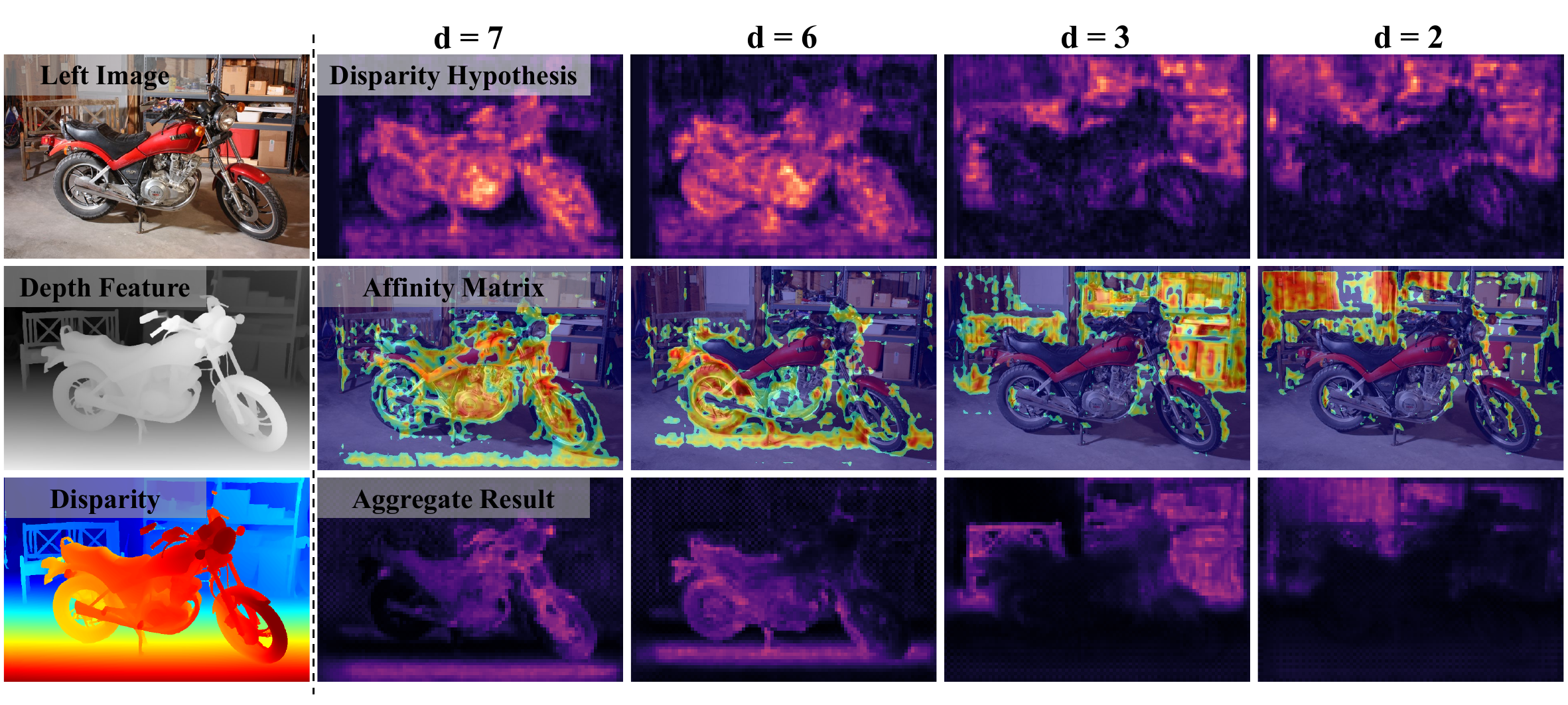} 
\caption{Visualization of affinity matrix across disparity hypotheses. The first row shows the initial cost volume features at different disparity hypotheses, which are fragile in unseen scenes and contain many mismatches. The second row visualizes the affinity matrices produced by our DDCA, which guide the aggregation network to focus on the most relevant regions for each disparity hypothesis. As a result, the aggregated cost volume (third row) preserves clearer object boundaries and more reliable geometric structures.}
\label{fig:affinitymatrix}
\end{figure*}

\begin{table*}[t]
\centering
\begin{tabular}{@{}l|ccccc|c|c@{}}
\toprule
\multirow{2}{*}{Variations} & Scene Flow & KITTI 2012 & KITTI 2015 & Middlebury & ETH3D & \multirow{2}{*}{\begin{tabular}[c]{@{}c@{}}Params.\\ (M)\end{tabular}} & \multirow{2}{*}{\begin{tabular}[c]{@{}c@{}}Time\\ (ms)\end{tabular}} \\
 & EPE & D1 & D1 & Bad 2.0 & Bad 1.0 &  &  \\ \midrule
3x3 Conv & 0.47 & 4.59 & 6.04 & 6.35 & 3.31 & 4.18 & 47 \\
3x3 DW+1x1 PW & 0.46 & 4.73 & 5.84 & 6.87 & 3.04 & 3.69 & 48 \\
1x1 Conv & 0.46 & 4.11 & 5.56 & 6.53 & 2.84 & 3.68 & 47 \\ \midrule
MoGe-2 (ViT-S) & 0.48 & 4.36 & 6.25 & 6.96 & 3.80 & 3.75 & 75 \\
Depth Anything V2 (ViT-S) & 0.46 & 4.11 & 5.56 & 6.53 & 2.84 & 3.68 & 47 \\ \bottomrule
\end{tabular}
\caption{Ablation study of SCF and DFE module.}.
\label{tab:scf}
\end{table*}

\begin{table}[t]
\centering
\small
{\setlength{\tabcolsep}{1mm}
\begin{tabular}{@{}cc|c|ccc@{}}
\toprule
Query & Key & Depth Prior & \begin{tabular}[c]{@{}c@{}}Scene Flow\\ EPE\end{tabular} & \begin{tabular}[c]{@{}c@{}}KITTI 2012\\ D1\end{tabular} & \begin{tabular}[c]{@{}c@{}}ETH3D\\ Bad 1.0\end{tabular} \\ \midrule
DH & DH & \ding{55} & 0.47 & 4.58 & 3.52 \\
DF & DH & \ding{51} & 0.46 & 4.33 & 3.04 \\
DH & DF & \ding{51} & 0.46 & 4.11 & 2.84 \\ \bottomrule
\end{tabular}
}
\caption{Ablation study on affinity matrix construction. DH: Disparity Hypothesis, DF: Depth Feature.}.
\label{tab:sddc_qk}
\end{table}


\subsection{Additional Comparisons with State-of-the-art}

\subsubsection{Scene Flow.}
As shown in Tab. \ref{tab:sceneflow}, our GGEV achieves the best performance among all published real-time methods on the Scene Flow test set.
We report results under different numbers of update iterations (2, 4, 6, and 8), allowing users to balance accuracy and runtime based on their application requirements.

\subsection{Additional Ablation Study}

\subsubsection{Convolution types of SCF.}
The upper part of Tab. \ref{tab:scf} presents an ablation study on SCF, where we investigate the impact of varying convolution kernel sizes and the use of depthwise separable convolution\cite{mobilenetv2}, a popular lightweight alternative.
Interestingly, despite its simplicity, the $1 \times 1$ convolution achieves superior performance compared to other alternatives.

\subsubsection{Generalization across MFMs in DFE.}
To evaluate the generalizability of GGEV to different monocular foundation models, we replace the depth feature extractor with MoGe-2 \cite{moge2}. As shown in the lower part of Tab.~\ref{tab:scf}, all GGEV variants consistently outperform RT-IGEV, demonstrating the versatility of our approach 
and its potential to benefit from future advances in monocular depth estimation.

\subsubsection{Effecitveness of Affinity Matrix Construction.} 
To validate the effectiveness of our affinity computation strategy in DDCA, we conduct an ablation study by varying the query and key inputs. As shown in Tab.~\ref{tab:sddc_qk}, dynamic convolution kernels with depth priors consistently yield superior performance. While constructing the affinity matrix solely from disparity hypotheses allows adaptive response to varying critical regions, the absence of structural guidance limits the ability to suppress mismatches. In contrast, incorporating depth features provides reliable structural cues that effectively guide the cost aggregation. The visualization of affinity matrix is shown in Fig. \ref{fig:affinitymatrix}.



\end{document}